\documentclass{article}

\usepackage{amsmath}
\usepackage{graphics}
\usepackage{amssymb}
\usepackage{natbib}
\usepackage{centernot}
\usepackage{hyperref}

\newcommand{\ve}[1]{\mathbf{{#1}}}  %
\newcommand{\bigCI}{\perp\mkern-10mu\perp}
\newcommand{\nbigCI}{\centernot{\bigCI}}

\newcommand{\w}{\mathbf{w}}
\newcommand{\ddim}{n} %

\newcommand{\s}{\mathbf{s}}
\newcommand{\x}{\mathbf{x}}

\newcommand{\n}{\mathbf{n}}
\renewcommand{\b}{\mathbf{b}}

\renewcommand{\u}{\mathbf{u}}
\newcommand{\f}{\mathbf{f}}
\newcommand{\g}{\mathbf{g}}
\newcommand{\R}{\mathbb{R}}
\newcommand{\W}{\mathbf{W}}
\newcommand{\B}{\mathbf{B}}
\newcommand{\U}{\mathbf{U}}
\newcommand{\myindex}{\text{$t$}} 
 
\newcommand{\A}{\mathbf{A}}
\newcommand{\J}{\mathbf{J}}

\newcommand{\thetab}{{\boldsymbol{\theta}}}

\newcommand{\ee}{\mathbf{e}}
\renewcommand{\t}{\tau}

\newtheorem{theorem}{Theorem}

\newtheorem{definition}{Definition}

\newtheorem{proposition}{Proposition}
\newcommand{\qed}{\ $\square$}

\newcommand{\diag}{\text{diag}}
\newcommand{\Ctwo}{\mathcal{C}^2}
\renewcommand{\b}{{\bf b}}

\renewcommand{\v}{{\bf v}}

\begin{document}

\title{Identifiability of latent-variable and structural-equation models: \\from linear to nonlinear}
\author{Aapo Hyv\"arinen,$^1$ Ilyes Khemakhem,$^2$ Ricardo Monti$^2$\\
$^{1)}$Dept of Computer Science, University of Helsinki, Finland\\
$^{2)}$Gatsby Computational Neuroscience Unit, UCL, UK}
\maketitle

\begin{abstract}
An old problem in multivariate statistics is that linear Gaussian models are often unidentifiable, i.e.\ the parameters cannot be uniquely estimated. In factor (component) analysis, an orthogonal rotation of the factors is unidentifiable, while in linear regression, the direction of effect  cannot be identified. For such linear models, non-Gaussianity of the (latent) variables has been shown to provide identifiability. In the case of factor analysis, this leads to independent component analysis, while in the case of the direction of effect, non-Gaussian versions of structural equation modelling solve the problem. More recently, we have shown how even general nonparametric nonlinear versions of such models can be estimated. Non-Gaussianity is not enough in this case, but assuming we have time series, or that the distributions are suitably modulated by some observed auxiliary variables, the models are identifiable. This paper reviews the identifiability theory for the linear and nonlinear cases, considering both factor analytic models and structural equation models.
\end{abstract}

%\keywords{Identifiability \and  independent component analysis \and  structural equation model \and  factor analysis \and  disentanglement \and  non-Gaussianity}

\textbf{Keywords:}\ {Identifiability ;  independent component analysis ;  structural equation model ;  factor analysis ;  disentanglement ;  non-Gaussianity}

\section{Introduction}

The goal of this paper is to provide a succinct and relatively self-contained exposition of the identifiability theory of a class of latent-variable models called independent component analysis, as well as of a class of structural-equation models. The theory has both linear and nonlinear versions, where ``nonlinear'' is to be taken in the sense of general (non-parametric) nonlinearities. The latent-variable models and structural-equation model are intimately related, and the identifiability theory of the former can be used to construct an identifiability theory of the latter. We focus on identifiability theory, and aim to explain the basic results on an approachable manner. Estimation methods and algorithms are given very little attention in this paper.

We start by motivating these different models in the rest of this section. The following sections are structured as follows. The notion of identifiability is defined in Section~\ref{id.sec}. The model of (linear) independent component analysis (ICA) is considered in Section~\ref{ica.sec}. Linear structural equation models (SEM)  are considered in Section~\ref{sem.sec}. Moving to nonlinear (non-parametric) models, Section~\ref{nica.sec} treats the identifiability of nonlinear ICA, and Section~\ref{nonsens.sec}, the identifiability of nonlinear SEM. Section~\ref{disc.sec} provides further discussion on the utility of identifiability, topics for  future research, and algorithms. Section~\ref{conc.sec} concludes the paper.

\subsection{Linear representation learning and factor analysis}

The problem of identifiability of latent variables was encountered already decades ago in the case of classical factor analysis, which forms the basis for all our developments. The basic model is as follows. Assume $s_i, i=1,\ldots,n$ are $n$ standardized uncorrelated Gaussian latent random variables. The covariance of $\s$ is thus identity, a property which is termed ``whiteness''. Assume $\A$ is an $m\times n$ matrix, and denote by $\n$ an $m$-dimensional vector of uncorrelated noise variables. We observe the $m$-dimensional random vector $\x$ which is a noisy linear mixture given by
\begin{equation} \label{fa}
\x=\A\s+\n
\end{equation}
The goal would be to recover the components in the vector $\s$, or at least the matrix $\A$. Many methods have been proposed \citep{Harman}, but fundamentally such factor analysis suffers from the indeterminacy of a ``factor rotation'', which is another way of saying that the factor analytic model is not identifiable for Gaussian factors. This is because any orthogonal transformation of the Gaussian latent vector, due to its whiteness, 
gives exactly the same distribution for the data, while giving quite different values of the latent variables, i.e.\ the features. This is a fundamental problem to which we will return more than once below. 

Thus, the parameter matrix $\A$ and the components $s_i$ cannot be uniquely recovered. 
Such unidentifiability is a problem since one important goal of fitting such models is to find the underlying structure of the data, or a useful \textit{representation}. If the representation a model gives is not unique or even well-defined, it is not possible to find the underlying structure of the data. It is therefore crucial to find models which are identifiable. The problem is particularly pertinent in the case where $m=n$, i.e.\ dimension reduction is not performed, which is our focus in this paper.\footnote{Dimension reduction in itself does not necessarily suffer from such lack of uniqueness if all we want to find is the right subspace. In the case of dimension reduction, factor analysis is actually closely related to principal component analysis (PCA). However, this whole paper is about the case where dimension reduction is \textit{not} performed; these two are very different problems.}

\begin{figure}
\begin{center}
{\textsf  Observed signals:\vspace*{1mm}}\\
\resizebox{\textwidth}{1cm}{\includegraphics{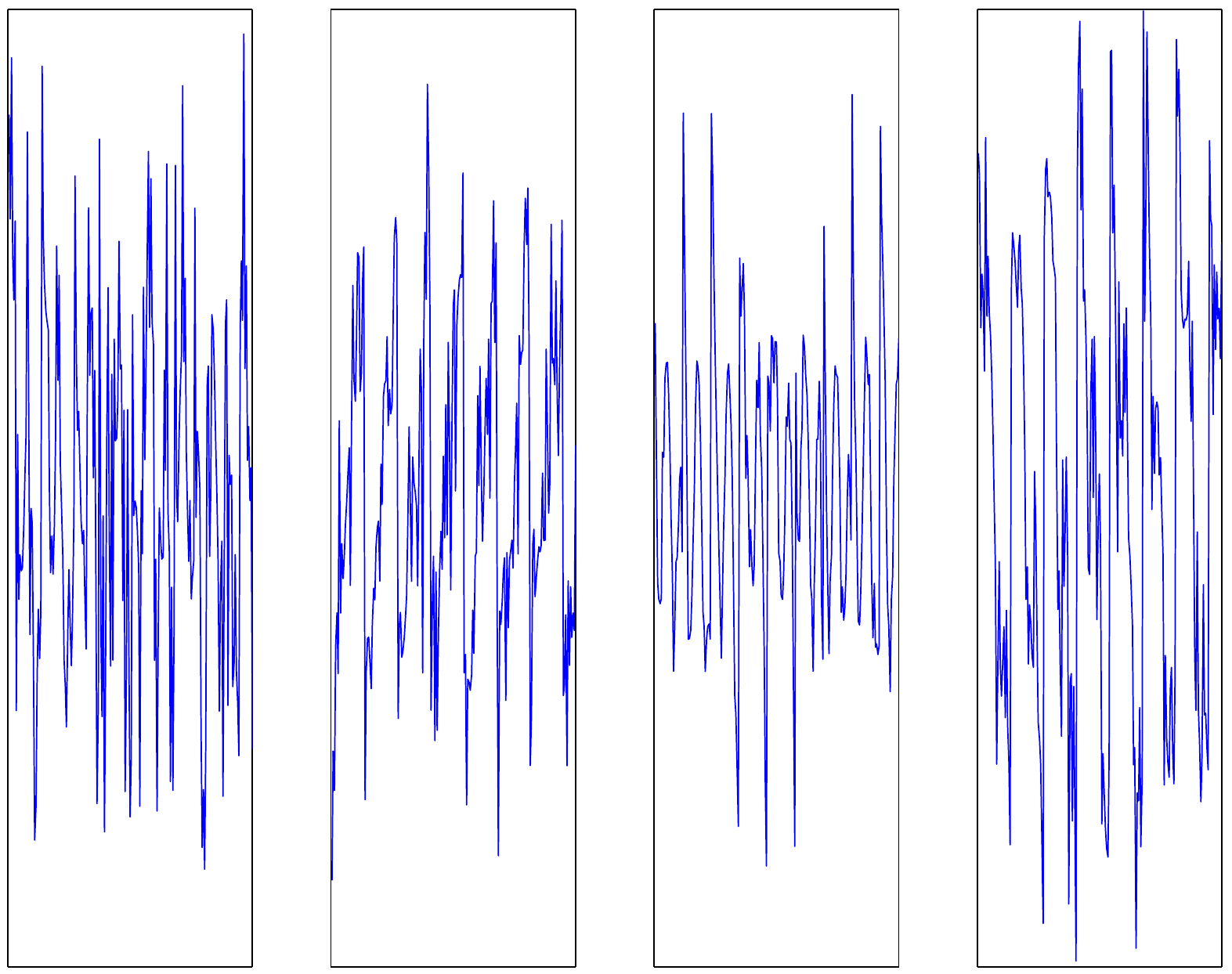}}\\
\vspace*{3mm}
{\textsf Independent components:\vspace*{1mm}}\\
\resizebox{\textwidth}{1cm}{\includegraphics{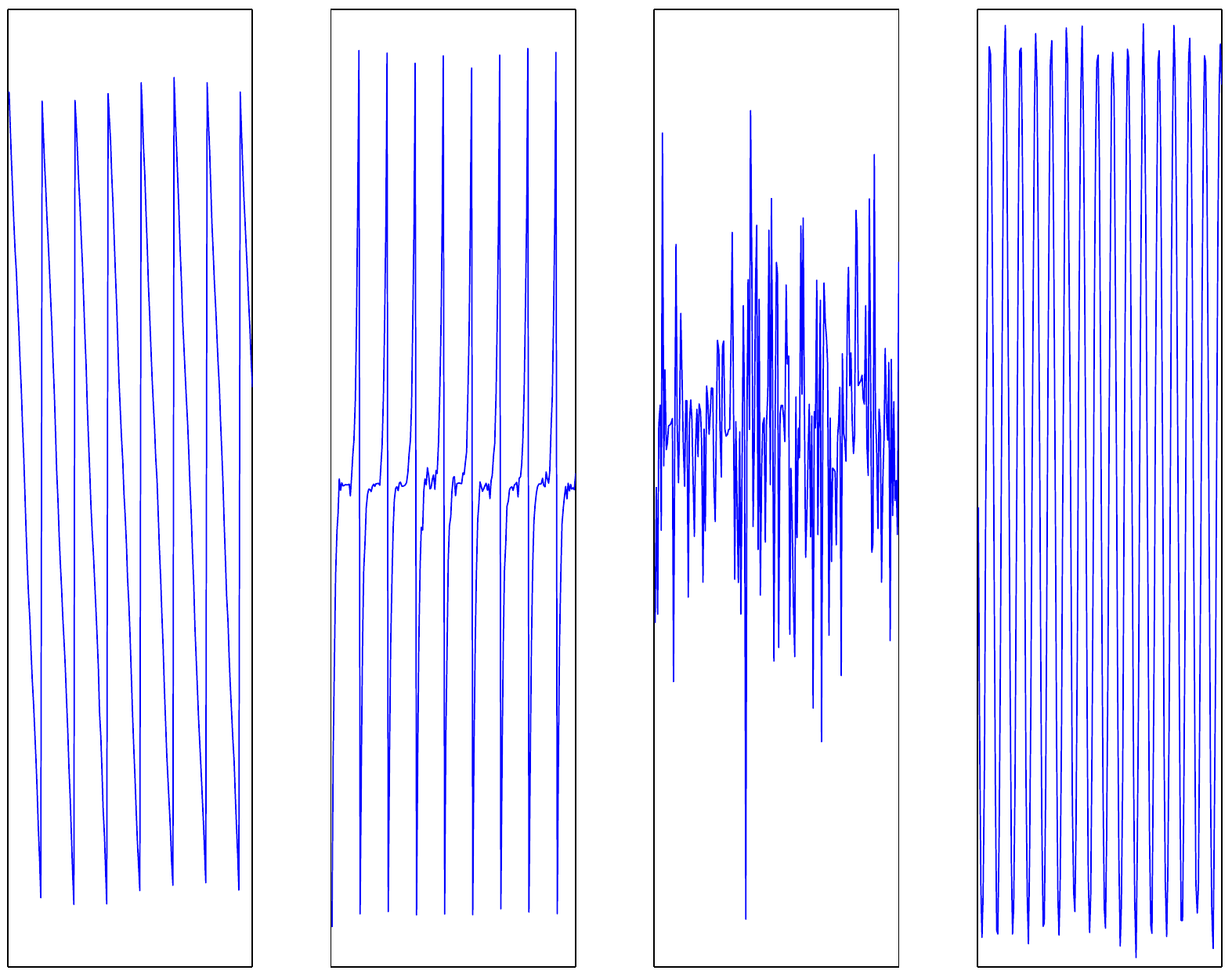}}
\end{center}
\caption{The basic idea of ICA. From the four measured signals shown in the upper row, ICA is able to recover the original source signals which were mixed together in the measurements, as shown in the bottom row. \label{icaillufig}}
\end{figure}

One practical motivation here is that we might want to separate signals from linear mixtures, and this should be done ``blindly'', i.e.\ using minimum information. This is illustrated in Fig.~\ref{icaillufig}. The four observed signals (to be taken as entries of a four-dimensional time series) are apparently unstructured or noisy. Can we find their underlying, hidden structure?

Fortunately, this problem of unidentifiability can be solved by independent component analysis (ICA) which assumes independent latent variables that are non-Gaussian. 
For example, applying ICA on the signal in Fig.~\ref{icaillufig}, we ``separate'' them into four original source signals which were seriously mixed in the observed data, as shown in the Figure. This kind of application of ICA is often called ``blind source separation''. If there were a factor rotation that remains undetermined, no such separation would be possible.

ICA will be reviewed in section  \ref{ica.sec} below and forms the basis of most of the theory reviewed in this paper.

\subsection{Nonlinear representation learning and disentanglement}

Machine learning has recently been profoundly transformed by deep learning, which essentially means learning (estimating) arbitrary non-parametric nonlinear functions from data. While the theory is quite well developed in the supervised case, consisting essentially of regression, the unsupervised case is much less developed. Unsupervised means that we only observe a potentially high-dimensional random vector $\x$, and there is no ``output'' or ``label'' or ``regressand'' defined (just like in factor analysis above). It is in fact widely appreciated that unsupervised, nonlinear  learning of representations or features is one of the biggest challenges facing machine learning at the moment. It is often referred to as ``disentanglement'', although this term is vague and not very well-defined.

It is often assumed that most satisfactory solution for unsupervised deep learning is based on estimation of probabilistic generative models, because probabilistic theory often gives optimal objectives for learning, and immediately enables probabilistic inference of various quantities, such as the latent variables. This brings the machine learning theory in close connection with statistical estimation. Finding a good representation can then be defined as recovering the original latent variables that were assumed to generate the data. 
Some of the most popular probabilistic methods for unsupervised deep learning are based on variational autoencoders (VAE; \citet{Kingma14}) and  generative adversarial networks (GAN;  \citet{Goodfellow14}). These methods have been successful in approximating the probability density function (pdf) of the data and generating new data points.

Unfortunately, in most models used in unsupervised deep learning, there has not been any proof that the original latent variables can be recovered. In fact, they tend to rely on transformations of latent variables that are  Gaussian and even white.  We thus recover the same serious problem of unidentifiability (factor rotation) as in the case of linear models already mentioned. 

Inspired by the theory of linear ICA, we might try to directly extend it to the nonlinear case. Such a basic framework would assume that the observed data is generated by an invertible nonlinear (non-parametric) transformation of non-Gaussian independent components, and we wish to recover them from observed data alone.

However, the extension of ICA to general nonlinear mixtures has proven very problematic. In particular, if the observed data $\x$ are obtained as i.i.d.\ samples, i.e.\ there is no temporal or similar structure in the data, the model is seriously unidentifiable \citep{Hyva99NN}. %
This is due to a further kind of unidentifiability that is specific to nonlinear models. 

Fortunately, a solution to non-identifiability in nonlinear ICA can be found by utilizing temporal structure in time series \citep{Harmeling03,Sprekeler14,Hyva16NIPS,Hyva17AISTATS} or similar ``auxiliary'' information \citep{Hyva19AISTATS,Khemakhem20iVAE}.
Section~\ref{nica.sec} reviews the identifiability theory for such nonlinear latent variable model, in particular nonlinear ICA.

\subsection{Causal discovery and structural equation models}

Causal discovery is another goal of statistical analysis. It may appear to be unrelated to latent variable models, but we will see below that there is actually a deep connection. Causal models play a fundamental role in modern scientific endeavor \citep{Spirtes2000, Pearl2009,peters2017elements}. 
While randomized control studies 
are the gold standard to study the effect of one variable on another, such an approach is unfeasible or unethical in many scenarios \citep{Spirtes2016}. 
Furthermore, big data sets publicly available on the internet often try to be generic and thus cannot be strongly based on specific interventions.
As such, it is both necessary and important to develop
\textit{causal discovery}
methods through which to uncover 
causal structure from (potentially large-scale) %
passively observed data.
Data collected without the 
explicit manipulation of certain variables is often termed
\textit{observational data}, in contrast 
to experimental data where certain variables are intervened upon, as in randomized controlled trials.

Before going to the general case, it is very useful to consider the linear bivariate case to understand the basic idea. The problem of causal discovery  then essentially means finding the ``direction of effect'' by choosing between two regression models; either
\begin{equation}
x_2=b x_1+e_2 
\end{equation}
which we might denote by $x_1\rightarrow x_2$, or
\begin{equation}
x_1=b x_2+e_1 
\end{equation}
which we might denote by $x_2\rightarrow x_1$.
It is well-known that if the variables are standardized \textit{Gaussian}, the situation is completely symmetric:
 The likelihood is equal for both models, and the variance explained is equal for both models (and so is the regression coefficient as already imposed above). Thus, we see again that for Gaussian variables we have a problem: the direction of effect is not identifiable. However, as will be seen below, it is identifiable for non-Gaussian variables.

\begin{figure}
\resizebox{0.29\textwidth}{!}{\includegraphics{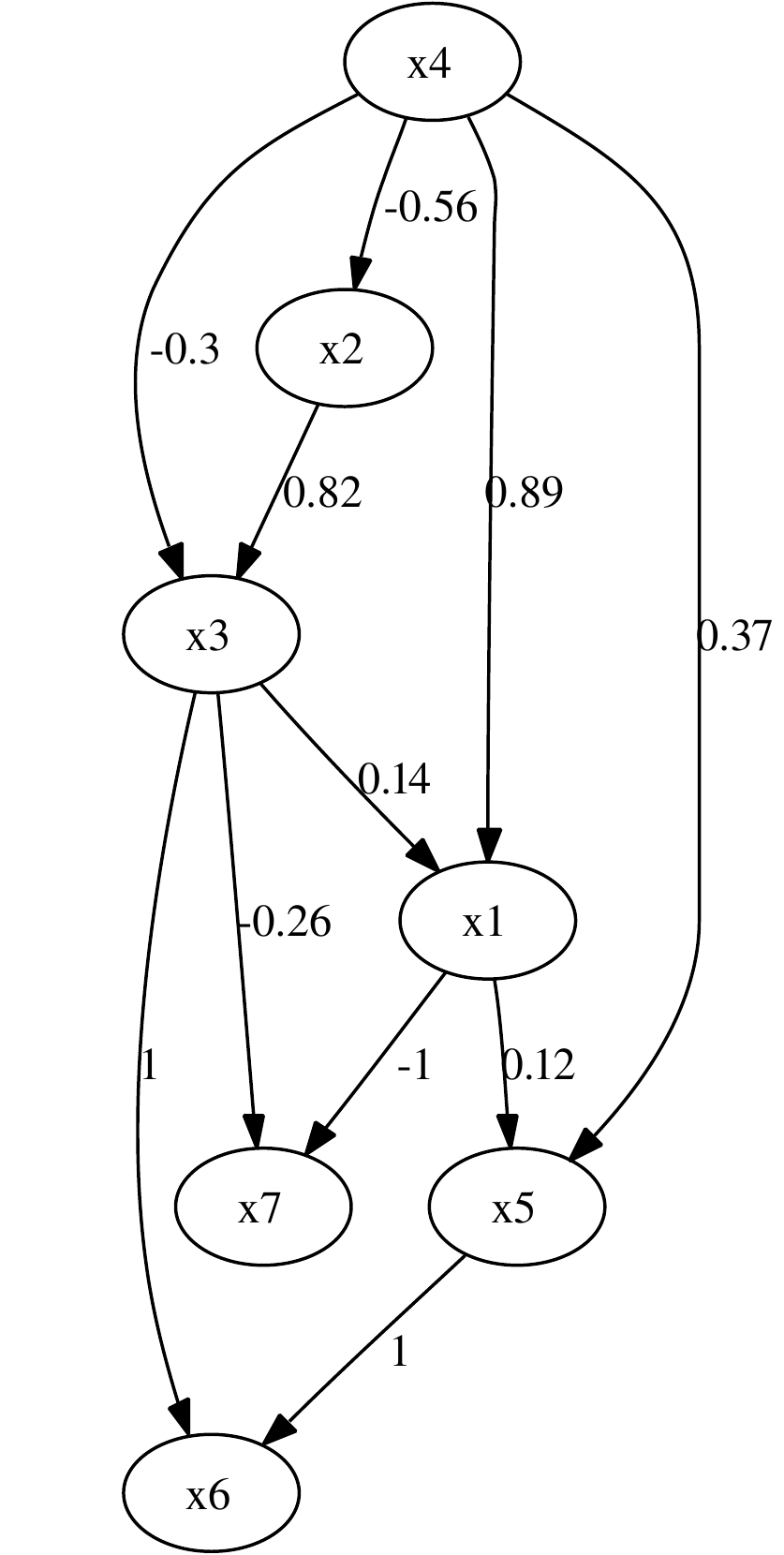}}
\caption{A SEM can be expressed by a directed graph (typically acyclic), where the arcs express causal influences, as well as statistical dependenceis. Here, the nodes have been ordered so that the influences all go from top to bottom. The disturbance or noise variables are not plotted here, since each observed variable simply has its own disturbance variable. \label{dag.fig}}
\end{figure}

In the general case, we use the framework of structural equation
models\footnote{Structural Equation Models (SEM) are sometimes referred to as Structural Causal Models
(SCM) or Functional Causal Models (FCM) in recent machine-learning literature.} (SEMs)
\citep{Bollenbook}. 
Fundamentally, SEMs define
a statistical model that describes the interactions of a set of observed variables
$\mathbf{x}=(x_1,\ldots,x_n)$ using a set of mutually independent disturbances
or noise variables $\mathbf{e}=(e_1, \ldots, e_n)$.
However, SEM is not only a statistical model of a probability distribution, but also a mathematical tool that can be used to
encapsulate causal knowledge \citep{Pearl2009}. SEMs are in this sense more powerful than latent variable models: not only do they describe the
set of all distributions, they can be used to perform interventions and answer
counterfactual queries by changing the noise distribution or the causal
mechanism in one or more of the equations~\eqref{eq:intro:sem}.
We leave a formal definition for later, and just illustrate the SEM by Fig.~\ref{dag.fig} which shows how the influences of the observed variables can be described by a graph, where the arcs show which $x_i$ causally influences which $x_j$. 

SEM are only useful for causal discovery if they define an \textit{identifiable} causal model.
In the case of a causal model, identifiability typically means that we can distinguish between cause and effect, or find the direction of effect as in the example above. In general, it means we can find the right ``causal ordering'' over the
variables $(x_1, \ldots, x_n)$ such that, loosely speaking, the earlier variables are the causes of the latter variables. %
Without identifiability, different causal orders would give rise to the same data, which would prevent any real scientific analysis of causality.

Although the problem of causal discovery may seem, at first sight, very different from estimating latent variable models, the deep result here is that we can in some cases reduce the estimation of a SEM to the estimation of a latent-variable model such as ICA. Thus, if we develop an identifiable  latent-variable model, we may be able to develop a corresponding SEM that is also identifiable. In fact, even the estimation methods can be largely shared.
Section~\ref{sem.sec} below will consider identifiability in  the linear case, while Section~\ref{nonsens.sec} will consider the nonlinear case.

\section{Definition of identifiability} \label{id.sec}

Now we shall proceed to a formal definition of identifiability. Identifiability is an important property of probabilistic models. For example, it seems futile to interpret the quantities estimated in the model, whether parameters or some further latent variables, if the model is not identifiable. We defer a more detailed discussion of the utility of identifiability to Section~\ref{disc.sec}, and simply give the definition and some examples here.

Consider a probabilistic model $\mathcal{P}$, defined as the set of distributions $ \{P_\thetab: \thetab \in \Theta\}$ with parameter $\thetab$ taking values in some set  $\Theta$, on a set of possible observations $\mathcal{X}$ (typically $\R^n$ in our case).
A model $\mathcal{P}$ is said to be \textit{identifiable} if
the mapping $\thetab \in \Theta \mapsto %
P_\thetab(\mathbf{x})%
$ is injective (i.e.\ one-to-one):
\begin{equation}
        \label{eq:intro:iden}
        \left( P_{\thetab_1}(\mathbf{x}) = P_{\thetab_2}(\mathbf{x}), \,
        \forall \mathbf{x} \in \mathcal{X}
        \right)
        \implies \thetab_1 = \thetab_2.
\end{equation}
In other words, if two parameters $\thetab_1$ and $\thetab_2$ generate the same
distribution over the set of observations $\mathcal{X}$, then they are
necessarily equal.
It is important to note that identifiability is a property of the probabilistic
model and not of any particular estimation method.

As a simple theoretical example, consider that we flipped a possibly biased coin $N$ times.
A coin toss only has two outcomes: heads with probability $\theta \in [0,1]$, or tails with probability $1 - \theta$. Let us define two different models for this data.
In the first case,
we use a Bernoulli model with parameter $\theta$ for the outcome of a coin flip; denote by $\textrm{Ber}_{\theta}(x)$ the probability of the coin flip given $x \in \{H, T\}$ which codes heads and tails.
This model is identifiable.
To see this, let $(\theta_1,\theta_2) $ be such that $\textrm{Ber}_{\theta_1}(x) = \textrm{Ber}_{\theta_2}(x)$  for $x \in \{H, T\}$.
Since $\textrm{Ber}_\theta(H) = \theta$, and this probability is directly given by the data in the limit of infinite coin flips, we conclude that $\theta_1 = \theta_2$ and that the model is identifiable.
On the other hand, as a counterexample, we can imagine something like a latent variable model: we do not observe a real coin, but rather the output of a computer simulation.
Tossing a coin in this case proceeds in two steps: we first draw a sample $z \sim \mathcal{N}(\mu, \sigma^2)$, i.e.\ a latent variable with a Gaussian
distribution with mean $\mu$ and variance $\sigma^2$;
then we assign heads to $x$ if $z \ge 0$ and tails otherwise.
This is effectively a latent variable model with parameters $(\mu, \sigma)$.
Crucially, elementary calculations show that the observed probabilities depend only on the ratio $\frac{\mu}{\sigma}$, which takes the same value for infinitely many pairs
$(\mu, \sigma)$, which means that this model is not identifiable.

As a most fundamental example for our purposes, consider the \textit{Gaussian factor analysis} in Eq.~(\ref{fa}), where the components or factors $s_i$ are Gaussian, uncorrelated, and have unit variance. This model is well-known not to be identifiable. For simplicity, consider the matrix $\A$ to be square and orthogonal--- denote it by $\U$ to highlight those assumptions---and ignore the noise $\n$.  An intuitive justification for unidentifiability would be that a Gaussian distribution is completely determined by covariances (and means). Now, the number of covariances is $\approx n^2/2$ due to symmetry, so we cannot solve for the $n^2$ parameters in the mixing matrix as we have ``more variables than equations''. More rigorously, the Gaussian distribution exhibits a rotational symmetry when the covariance matrix is identity (as is the case for the components here). The pdf is given by the probability transformation formula (transforming $\x$ back to $\s$):
\begin{equation}
  p(\x)=\frac{1}{(2\pi)^{n/2}}\exp(\frac{1}{2}\|\U^T\x\|^2)|\det \U^T| =\frac{1}{(2\pi)^{n/2}}\exp(\frac{1}{2}\|\x\|^2)
\end{equation}
where the last equality comes from the orthogonality of $\U$. Now, we see that the pdf of $\x$ does not depend on $\U$ at all. Thus, $\U$ cannot be identifiable. In practice, we could rotate the factors $\s$ by any orthogonal matrix, and compensate for that by rotating the columns of $\U$ by the inverse, and the data distribution would stay the same, which is why this is called the ``factor rotation indeterminacy''. %

The basic definition of identifiability talks thus about identifiability of parameters. Sometimes, by slight abuse of terminology, we also talk about \textit{identifiability of the latent variables}, but it is not quite clear how that should be defined. If we consider the factor analysis model in Eq.~(\ref{fa}) without noise, knowing the matrix $\A$ will immediately give us the components $\s$ by using the (pseudo)inverse of $\A$ (assuming $m\geq n$ as is typical). Thus, identifiability of $\A$ implies, in a non-rigorous sense, identifiability of $\s$. On the other hand, if there actually is noise in the model, knowing $\A$ will not give us the components $\s$ since the noise cannot be completely removed. In this case, identifiability could be defined in the sense that the posterior $p(\s|\x,\A)$ can be recovered in a suitable sense \citep{Khemakhem20iVAE}. However, we leave such a definition aside here, and focus on identifiability of the parameters.

It should also be noted that the strict definition of identifiability in Eq.~(\ref{eq:intro:iden}) may be limiting in some cases. In practical scenarios, we may want to introduce identifiability with slightly relaxed definitions, such as identifiability of parameters up to an equivalence class. For example, independent components are only identifiable up to arbitrary scaling  of the components, but this is usually not considered a problem.

\section{Linear Independent Component Analysis} \label{ica.sec}

ICA is a statistical latent variable model which is very closely related to the classic factor analysis model reviewed above. The basic idea is that assuming the components to be non-Gaussian breaks the rotational symmetry just described and leads to an identifiable model. In particular, for non-Gaussian data, higher-order moments give more information than that contained in the covariances \citep{Hyva00NN,Hyvabook}. The identifiability of the basic ICA model is well-known since \citet{Comon94}, and in fact was proved in the 1950's by Darmois and Skitovich.

\subsection{Definition and identifiability}

The basic model is as follows \citep{Comon94,Jutten91,Hyva00NN,Hyvabook}. Assume $s_i, i=1,\ldots,n$ are $n$ independent, non-Gaussian latent random variables. Assume $\A$ is an invertible $n\times n$ matrix. We observe the random vector $\x$ which is a linear mixture given by
\begin{equation} \label{ica}
\x=\A\s
\end{equation}
We assume that the means of the $s_i$ are centered to zero and that their variances are finite and normalized to unity. Note the differences to the factor analysis model:
\begin{enumerate}
\item The components are assumed to be \textit{non-Gaussian}. This is the fundamental difference which sets the model apart from classical factor analysis. (While the first letter in ``ICA'' emphasizes independence, the factors in classical Gaussian factor analysis are independent as well; the terminology is slightly misleading regarding this difference).
\item The number of components (factors) is equal to the number of observed variables. This is basically assumed for the sake of mathematical simplicity: the matrix $\A$ is then uniquely invertible, but the assumption could be relaxed. At the same time, it emphasizes the fact that in the context of ICA we are not so interested in dimension reduction but on finding the original components. In practice, the dimension is often reduced by principal component analysis (PCA) before application of ICA. In terms of the classical terminology of factor analysis, ICA can then be seen as a ``factor rotation'', considering the principal components as estimates of factors.
\item There is no noise term. This is partly justified by the large number of components: some of them can be noise whereas others are more interesting components. Since we do not reduce the dimension of the data, all variance of the data is ``explained'' by the components anyway.
\end{enumerate}

It is really the non-Gaussianity that fundamentally distinguishes the ICA model from classical factor analysis, and enables identifiability.
The importance of non-Gaussianity is further emphasized by the theory of ICA estimation, which provides also an intuitive proof of identifiability as follows.  One of the most fundamental theorems in ICA says that we can estimate ICA by finding an invertible transformation that maximizes the non-Gaussianity of the components. In fact, each component corresponds to a maximum (over some parameters $w_i$) of non-Gaussianity of a linear combination $\sum_i w_i x_i$. This is because, loosely speaking, by the Central Limit Theorem, a sum of independent random variables is more Gaussian than any of the original random variables (strictly speaking this holds if the random variables have the same distribution). The linear combination $\sum_i w_i x_i$ is a linear combination of the $s_i$, and thus its non-Gaussianity is maximized when it is actually equal to one of the $s_i$. See \citet{Hyva00NN,Hyvabook} for details.

Note two rather trivial indeterminacies, i.e.\ unidentifiable aspects of the ICA model. First, the ordering of the components is not identifiable, or even defined by the model. Second, each component can only be estimated up to linear scaling (and sign), since if a component is multiplied by a scalar constant and the corresponding column of $\A$ is divided by that constant, the data distribution stays the same. This indeterminacy is partly removed when the variances of the components are conventionally defined to be equal to unity, which also means that $\s$ is white, but this is a mere convention and the real variance of the components is unidentifiable.

\subsection{A simple identifiability proof}

Next, we provide a simple rigorous identifiability proof for ICA. %
The price to pay for its simplicity is some reduction in generality compared to the celebrated Darmois-Skitovich theory. %
In our simple case, the identifiability theorem takes then the following form:
\begin{theorem}
  Assume the independent components have finite variance, and the log-pdf's of the independent components have continuous second derivatives. If the variables $x_i,i=1,\ldots,n$ in Eq.~(\ref{ica}) are mutually independent, $\A$ has exactly one non-zero entry in each row and each column.
\end{theorem}

\textit{Proof:}\footnote{Alternatively, we could do the same proof in the Fourier domain, i.e.\ using characteristic functions $\hat{p}$.  Then,
 $p(\x)$ and $p_i(s_i)$ will be replaced by the characteristic functions, and the Jacobian disappers in the first equations. 
Thus, we would replace the assumption of smooth pdf's by the assumption of continuous second derivatives of the characteristic functions of the $p_i$, denoted by $\hat{p_i}$. Such an assumption is related to the moment structure of the components: it is just slightly more restrictive than assuming finite variances for the components. The whole proof is valid for characteristic functions with minimal changes. Thus we get a more general if a bit more complicated proof.}
As typical in ICA theory, we can assume without restriction of generality that $\A$ is orthogonal. This is because $\s$ is white 
and we also can whiten $\x$ as preprocessing, which implies the orthogonality of $\A$ since the covariance of $\x$ is equal to $\A\A^T$.

The pdf of $\x$ is obtained by the formula for transformation of probability density as
\begin{equation}
p(\x)=\prod_{i=1}^n p_i(\sum_{j=1}^n a_{ji} x_j)|\det \A^T|
\end{equation}
where $p_i$ denotes the pdf of the $i$-th component. Taking logarithms, and noting that the determinant of an orthogonal matrix is equal to $\pm 1$, we get
\begin{equation}
\log p(\x)=\sum_{i=1}^n \log p_i(\sum_{j=1}^n a_{ji} x_j)
\end{equation}
Next, since the $x_i$ are assumed mutually independent,
$
\log p(\x)=\sum_{i=1}^n f_i(x_i)
$ for some functions $f_i$.
This implies that any second-order cross-derivative must be zero: %
\begin{equation}
\frac{\partial^2 \log p(\x)}{\partial x_k \partial x_l}= \sum_{i=1}^n  a_{ki} a_{li} (\log p_i)''(\sum_{j=1}^n a_{ji} x_j) =0, \; \; \text{ for all }k\neq l
\end{equation}
This set of equations can be collected together and expressed in matrix form as
\begin{equation} \label{main}
\A^T \diag_i[ (\log p_i)''(\sum_{j=1}^n a_{ji} x_j) ] \A = \diag_i[c_i(\x;\A)]
\end{equation}
where the $c_i(\x;\A)$ are some unknown scalar-valued functions of $\x$ and $\A$ corresponding to the case $k=l$. This equation must hold for all $\x$. 

For Gaussian densities (and no others), $\log p_i$ is quadratic, and its second derivative is constant. We have assumed non-Gaussianity, so $(\log p_i)''$ is not constant. By continuity, $(\log p_i)''$ takes in fact an infinity of different values, since it takes all the values in some interval of the real line. This implies we can find for each $i$ a point $y_i$ such that the entries $d_i=(\log p_i)''(y_i)$ are all distinct, i.e.\ $d_j\neq d_k$ for $j\neq k$. (This is possible even if one of the components is Gaussian.) By the invertibility of $\A$, we can find corresponding $\x$ such that $y_i= \sum_{j=1}^n a_{ji} x_j$ for all $i$. In the following, we consider (\ref{main}) with $\x$ fixed to such value, so the diagonal entries $d_i$ and $c_i$ are fixed.

Now, the equation on the LHS of (\ref{main}) is actually an eigen-value decomposition (EVD), since $\A$ is orthogonal. Importantly, the diagonal entries (eigenvalues) are distinct, which implies by a well-known results in linear algebra that the EVD is unique up to ordering of the eigenvalues and the signs of the eigenvectors. The RHS can be interpreted as an EVD as well, with eigenvectors such that each row and column has exactly one non-zero entry. Since the ``eigenvectors'' of the EVD on both sides must match up to permutation,  the eigenvectors of $\A$ must have the same property for non-zero entries.\qed

Note that if the components were Gaussian, the diagonal matrix on the left-hand-side of (\ref{main}) would be equal to identity, and thus the whole LHS is equal to identity for any $\A$. Then, any orthogonal $\A$ fulfills (\ref{main}), since it is easy to show that the right-hand-side must necessarily be identity as well. This shows how Gaussian variables are not allowed. In fact, the uniqueness of the eigen-value decomposition only holds for distinct eigenvalues.

Another point to note is that the matrix $\A$ is not shown to be identity. The form that $\A$ takes enables a permutation of the components, as well as multiplying each by a scalar constant. These are fundamental indeterminacies that cannot be solved in any model similar to factor analysis, as already pointed out above.

\if0
We start by computing the characteristic function of $\x$ as
\begin{equation}
  \hat{p}_\x(\u)=
  E\{\exp(\sqrt{-1}\u^T\A\s\}=  \prod_iE\{\exp(\sqrt{-1}(\sum_j u_j a_{ji})s_i\}=\prod_i \hat{p}_i(\sum_j u_j a_{ji})
\end{equation}
and for logarithms we have
$
\log \hat{p}(\u)=\sum_i \log \hat{p}_i(\sum_j u_j a_{ji})
$
Now, if the $x_i$ are independent, this must be a sum of some functions of each $u_i$ only:
$
\log \hat{p}_\x(\u)=\sum_{i=1}^n \tilde{f}_i(u_i)
$.
This leads to a similar condition on partial derivatives
\begin{equation}
\frac{\partial^2 \log \hat{p}_\x(\u)}{\partial u_k \partial u_l}= \sum_{i=1}^n  a_{ki} a_{li} (\log \hat{p}_i)''(\sum_{j=1}^n a_{ij} u_j) =0, \; \; \text{ for all }k\neq l
\end{equation}
Again, the logarithm of the characteristic function is non-quadratic for non-Gaussian densities, so the rest of the proof follows identically, replacing $\W$ by $\A^T$ and $p_i$ by $\hat{p_i}$. Such a proof technique  would actually be quite close to what is often used for the Darmois-Skitovich theorem.
\fi

%

\subsection{Alternative approaches} \label{altica.sec}

The literature of blind source separation actually considers non-Gaussianity as only one, if the most important, principle that enables estimation of a linear mixing model. In the case of time series data, identifiability is enabled by two very different kinds statistical structure: autocorrelations of the components \citep{Tong91,Belo97} 
or their non-stationarity \citep{Matsuoka95,Pham01}. Together with non-Gaussianity, these constitute what \citet{Cardoso01} called the ``three easy routes to [linear] ICA''.  We shall consider these principles in more details in the case of nonlinear ICA below.

Yet another approach is possible in the case where the data is non-negative, and in particular so that the data has a concentration to small positive values. This leads to the model originally called positive matrix factorization \citep{Paatero94} and popularized under the heading of non-negative matrix factorization (NMF) by \citet{Lee99Nature}.
Its identifiability has been analyzed by \citet{Donoho04}. Combinations of NMF and ICA are considered by \citet{Plumbley03,Hoyer04JMLR}. Related to this, \citet{Hyttinen22} consider the case where the observations are binary.

We further note that great similarity of the ICA model with dictionary learning and sparse coding. The similarity was well-known in early work of those methods \citep{Olshausen97} but seems to have been forgotten recently. The only real differences between ICA and a probabilistic formulation of dictionary learning is that in dictionary learning, the number of components is large ($n>m$), and there is noise. Some identifiability results for such cases are considered by \citet{Eriksson04}.

\section{Linear Structural Equation Model} \label{sem.sec}

Next, we consider a linear SEM, which we use here as a fundamental framework for causal discovery. We will see how its identifiability can be proven based on the theory of linear ICA.

\subsection{Definition of model}

A linear SEM  consists of a collection of equations of the form
\begin{equation}
        \label{eq:linearsem}
        x_i = \sum_{j=1}^\ddim b_{ij} x_j + e_i, \quad i=1,\ldots,\ddim,
      \end{equation}
      where the $x_i$ are the observed variables, and the $e_i$ are latent variables called the disturbances, external influences, or simply noise variables.
The idea is that the variable $x_i$ is caused by those $x_j$ for which the coefficient $b_{ij}$ is non-zero.

An SEM is often associated with a directed acyclic graph (DAG) $\mathcal{G}$ called the \textit{causal graph}. Each node of $\mathcal{G}$ corresponds to
an observed variable $x_i$, and there is an edge from $x_j$ to $x_i$ iff $b_{ij}$ is non-zero.
We have here actually made the assumption of acyclicity very typical in this context. It means that when the matrix $\B$ is interpreted as a connection matrix of a (weighted) graph, the ensuing graph has no cycles: one cannot start at a node and follow the edges so that one comes back to the starting node.
An interesting property of a DAG is that the nodes (variables) can be ordered so that all the connections go ``forward'' in that ordering; this is called a ``causal order'', but it is not necessarily unique. An illustration of such a DAG expressing a SEM, visually ordered according to the causal ordering,  was given in Figure~\ref{dag.fig}.

The problem is now to estimate the parameters $b_{ij}$ based on observations of $\x$. It is well-known that the problem is in general ill-posed. In particular, for Gaussian data the model is unidentifiable, as shown above for the case of two variables.
One wide-spread solution is to use prior knowledge to fix most of the $b_{ij}$ to zero; for example, randomized controlled trials might be available to provide that knowledge. However, using such prior knowledge, not to mention interventions, is in strong contradiction with the aim of \textit{discovery} which is central to us.

\subsection{Identifiability}

The key to identifiability of the SEM is its close relation to ICA. This intimate relation of latent-variable models and SEMs is a central theme of this paper, and we will later see how it applies even in the nonlinear case.

Denote by $\B$ a matrix which collects all the coefficients $b_{ij}$. We can express the SEM as 
 \begin{equation} 
   \x= \mathbf{B}\x+\ee
 \end{equation}
where the vector $\ee$ collects the external influences. Now, by elementary linear algebra this implies 
 \begin{equation} \label{semica}
 \x=(\mathbf{I}-\mathbf{B})^{-1}\ee
 \end{equation}
 In other words, the SEM implies that the data follows a latent-variable model. In fact, this is nothing else than an ICA model with mixing matrix $\A=(\mathbf{I}-\mathbf{B})^{-1}$, under suitable assumptions. In particular, assume that the $e_i$ are \textit{mutually independent and non-Gaussian}. Then, the model in Eq.~(\ref{semica}) is exactly an ICA model, and has by definition the same number of components (corresponding to disturbances) as observed variables.

 Thus it would seem that we can estimate an ICA model of $\x$, and transform the obtained $\A$ back to $\B$ by simply $\B=\mathbf{I}-\mathbf{A}^{-1}$. However, there is one serious complication: ICA does not estimate order of the components, $e_i$. As pointed out in Section~\ref{ica.sec},  there is no ordering inherently defined between the components. This is in stark contrast to the SEM, where we know that $e_i$ is the external influence of the variable $x_i$ (for the same index $i$). Thus, we need to find the right ordering of the independent components, which implies an ordering of the columns of the mixing matrix $\A$, before we can transform back to $\B$.

 The recovery of the correct ordering is not possible in general. However, as mentioned above, in the theory of SEM and causal discovery the assumption of a directed \textit{acyclic}  graph (DAG) is often made.   Based on the acyclicity assumption, the right ordering of the components can be found. To see how this is possible, consider the 2D case. Assume we have estimated the inverse of the mixing matrix $\W=\A^{-1}$, for some arbitrary ordering of the components. Based on Eq.~(\ref{semica}), we see that we should have $\W=\mathbf{I}-\B$. The acyclicity of $\B$ means in this 2D case that it has exactly one non-zero entry, by definition in the off-diagonal (unless the graph is degenerate and $\B$ is all zeros). Thus, $\W$ has exactly one zero entry. Denoting an arbitrary non-zero entry by $*$, the real $\W$ could, for example, be of the form
 \begin{equation}
   \W=\begin{pmatrix} 1 & * \\ 0 & 1 \end{pmatrix}
\end{equation}
Now, if the rows of $\W$ are switched to the wrong order in the estimation due to the indeterminacy of the order of components, it is easy to see that the zero goes to the diagonal. This is a contradiction since the diagonal entries are all one by definition (even allowing for any rescaling). Thus, among the two different orderings, it is possible to find the right one based on acyclicity. The general case is explained by \citet{Shimizu06JMLR}; note that acyclicity is sufficient but not necessary \citep{Lacerda08}.

 A smaller detail is that the normalization of the mixing matrix in ICA and SEM is different: In ICA, the variances of the components are typically defined to be unity, while in SEM, a related normalization is obtained by the fact that $\mathbf{I}-\B$ has all ones in the diagonal. However, this is just a normalization convention that has little implication for identifiability.

 Thus, we see that the SEM is identifiable under the assumptions of independence and non-Gaussianity (like in ICA), together with the new assumption of acyclicity. The resulting model is called LiNGAM for Linear Non-Gaussian Acyclic Model. We refer the reader to \citet{Shimizu06JMLR} for details on this basic model, and \citet{Shimizu2014} for a longer treatment with some more recent developments. 

Regarding estimation of LiNGAM, it is also possible to develop a very explicit interpretation of the resulting SEM estimation in terms of maximization of non-Gaussianity \citep{Hyva13JMLR}, just like in the case of ICA estimation. 

\subsection{Alternative approaches}
 
Alternative identifiable SEM frameworks have been proposed, e.g.\ by \citep{peters2014identifiability,jakobsen2022structure}. These remove the non-Gaussianity assumption but introduce alternative restrictions more inspired by causal inference literature. One particularly important problem with causal discovery is that we may not observe all the relevant variables; discovery in the presence of such hidden ``confounders'' is considered, e.g.\ by \citet{Hoyer08,Tashiro14NC}. A combination of source separation and LiNGAM was proposed by \citet{Monti18UAI}.
Finally, we note that the alternative methods for estimating a linear mixing model, reviewed in Subsection~\ref{altica.sec} above, could also be used to estimate a linear SEM.
In fact, \citet{Zhang10UAI} use a related idea to develop another combination of source separation and causal discovery.

\section{Nonlinear Independent Component Analysis} \label{nica.sec}

Nonlinear ICA is a fundamental problem in unsupervised learning which has attracted a considerable amount of attention recently. It promises a principled approach to representation learning and ``disentanglement'', in particular using deep neural networks. Nonlinear ICA attempts to find nonlinear components in multidimensional data by generalizing the linear ICA framework in Section~\ref{ica.sec}. The essential difference to most methods for unsupervised representation learning is that the approach starts by defining a generative model in which the original latent variables can be recovered, i.e.\ the model is identifiable by design.

\subsection{Problems in identifiability}

We start by introducing a simple model for nonlinear ICA which turns out to be unidentifiable. Denote, as above, an observed $n$-dimensional random vector by $\x=(x_1,\ldots,x_n)$. We assume it is generated using  $n$ independent latent variables called independent components, $s_i$. A straightforward definition of the nonlinear ICA problem is to assume that the observed data is an arbitrary (but smooth and invertible) transformation $\f$ of the latent variables $\s=(s_1,\ldots,s_n)$ as
\begin{equation} \label{genmix}
\x=\f(\s)
\end{equation}
The goal is then to recover the inverse function $\f^{-1}$ as well as the independent components $s_i$ based on observations of $\x$ alone.

Research in nonlinear ICA has been hampered by the fact that such simple approaches to nonlinear ICA are not identifiable, in stark contrast to the linear ICA case. 
To put it simply, for any $x_1,x_2$, one can always find a
function $g(x_1,x_2)$ which is independent of $x_1$.
Darmois provided such a construction back in 1952, thus showing the ``impossibility'' of  nonlinear ICA. He simply used the conditional cumulative distribution function 
  \begin{equation}
    g(\xi_1,\xi_2)= P(x_2<\xi_2 |x_1=\xi_1)
  \end{equation}
  to  construct a new variable $z=g(x_1,x_2)$ which turns out to be independent of $x_1$.
  (A slight generalization was provided by \citet{Hyva99NN}, and a slight variation by \citet{Locatello2019}.)
The problem is that we can apply this equally well after making an initial transformation
$\tilde{x}_1=h(x_1,x_2)$, so we see that \textit{any} such
transformation $\tilde{x}_1$ could be considered an independent
component, since we can always find a decomposition of the data as $\tilde{x}_1$ and the variable defined by the construction above. The marginal distributions of the two variables can further be easily transformed to anything we like, so we can, in particular, reproduce the \textit{distributions} of the original independent components exactly without recovering the \textit{components} themselves. (This logic is a bit different from  the definition of identifiability given above, since here we consider the distributions of the latent variables, but it is equivalent.)
Thus, it is clear that independence is not a strong enough assumption to enable identifiability, whatever assumptions (e.g.\ non-Gaussianity) we may make on the distributions of the independent components $s_i$.

The unidentifiability is of course a major problem in practice since most of the utility of ICA rests on the fact that the model is identifiable, or in alternative terminology, the ``sources can be separated''. We should also note that curiously, many unsupervised deep learning models, such as VAE and GAN, assume a latent-variable model where the $s_i$ are even Gaussian, thus recreating even the classical Gaussian factor rotation problem and making the unidentifiability even worse.

\subsection{Identifiable model definition}

However, it should be noted that Darmois's counterexample above assumes that the data is sampled i.i.d. (independently and with identical distributions) from a random vector; in particular, the observations (data points) are statistically independent of each other (which is not to be confused with the independence of the components).
A promising direction is to relax the assumption of i.i.d.\ sampling. A fundamental case  is to consider time series, and the \textit{temporal structure} of independent components. Thus, we assume
\begin{equation} \label{genmixtemp}
\x(t)=\f(\s(t))
\end{equation}
where $t$ is the time index.
As a first attempt, we can assume that the sources $s_i(t)$ have non-zero \textit{autocorrelations}, which has a long history in the linear case \citep{Tong91,Belo97}.
In machine learning literature, such models have been widely used under the heading of ``temporal coherence'', ``temporal stability'', or ``slowness'' of the features. 
\citet{Harmeling03} proposed that we could try to find nonlinear transformations which are maximally uncorrelated even over time lags and after nonlinear scalar transformations. \citet{Sprekeler14} showed that a closely related method enables separation of sources if they all have distinct autocorrelations functions. \citet{Sprekeler14} thus constitutes probably the first identifiability proof for nonlinear ICA with general nonlinearities; however, it suffers from the restrictive condition that the sources must have different statistical properties, which is rather unrealistic in many cases. Note that just like the linear scaling is unidentifiable in linear ICA, a nonlinear scaling by a monotonic function is fundamentally unidentifiable in nonlinear ICA; in both cases, the global sign of each component as well as their ordering are unidentifiable as well.

\citet{Hyva17AISTATS} proposed a rigorous and general treatment of identifiability for components which are stationary time series, the components being independent from each other, but each component having \textit{temporal dependencies}. Assuming the components $s_i(t)$ have sufficient temporal dependencies and sufficient non-Gaussianity according to certain technical definitions, the model was proven to be identifiable, up to monotonic pointwise transformations of the components. The proof was refined by \citet{Halva21}, partly based on \citet{schell2023nonlinear}; an alternative approach was proposed by \citet{klindt2020towards}.

Intuitively speaking,  temporal dependencies can provide much more information since now the components are independent over any lags, i.e.\ $s_i(t)$ and $s_j(t-\tau)$ are independent for any lag $\tau$. Such independence can then be imposed on any estimates of the components. This provides many more constraints compare to the i.i.d.\ case considered in the Darmois construction given above \citep{schell2023nonlinear}. Therefore, it is intuitively plausible that the model becomes identifiable.

Another form of temporal structure that has been previously used in the case of linear blind source separation is \textit{nonstationarity} \citep{Matsuoka95,Pham01}, in particular nonstationarity of variance.  Such nonstationarity of variances seems to be prominent in many kinds of real data, for example in EEG and MEG  \citep{Brookes11}, natural video \citep{Hyvanisbook}, and closely related to changes in volatility in financial time series. 
This principle was extended to the nonlinear case by \citet{Hyva16NIPS}; \citet{Khemakhem20iVAE} gives the strongest identifiability results based on this principle so far. They assumed a piecewise nonstationary model where the components in each segment follow an exponential family, while the parameters of the exponential family change from one segment to another. It was further assumed that the segmentation is known. Then the identifiability of the model was proven under technical assumptions that guarantee the nonstationarity is strong enough. While the assumptions here look quite restrictive, this result can be extended in various ways as will be seen below. Note that in this theory, the components can usually only be recovered up to a nonlinear, pointwise but not necessarily monotonic,  function of the components; for example, the only squares of the components can be recovered in the simplest models.

Again, we can give the identifiability results a simple intuitive justification. By the basic assumption of independence, the components are independent at any time point, i.e.\ $s_i(t)$ and $s_j(t)$ must be independent for any $t$. Crucially, nonstationarity implies that the distributions inside the segments are different; thus, the number of independence constraints is basically multiplied by the number of segments. This way we get many more constraints compared to the i.i.d.\ case, which was considered in the Darmois construction given above. 

\begin{figure}
\parbox{4.5cm}{ \begin{center}
\resizebox{3.5cm}{0.8cm}{\includegraphics{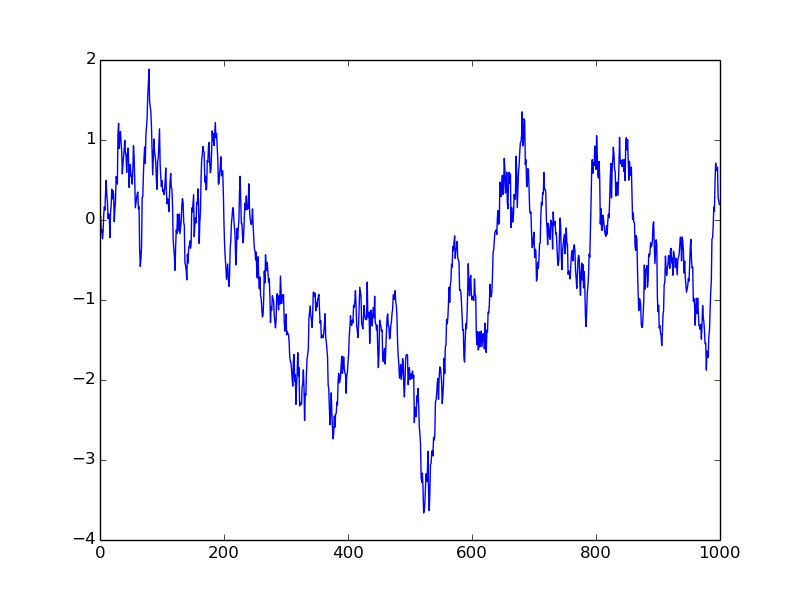}}
\resizebox{3.5cm}{0.8cm}{\includegraphics{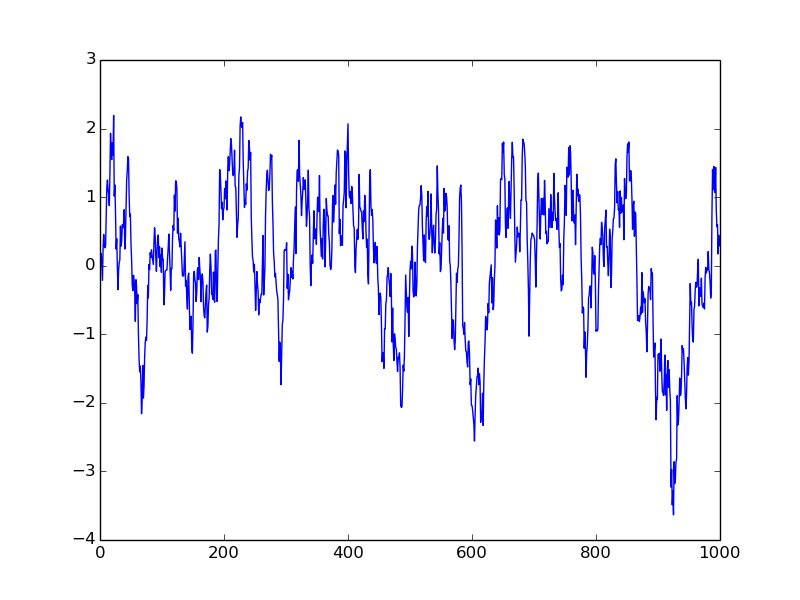}}
\resizebox{3.5cm}{0.8cm}{\includegraphics{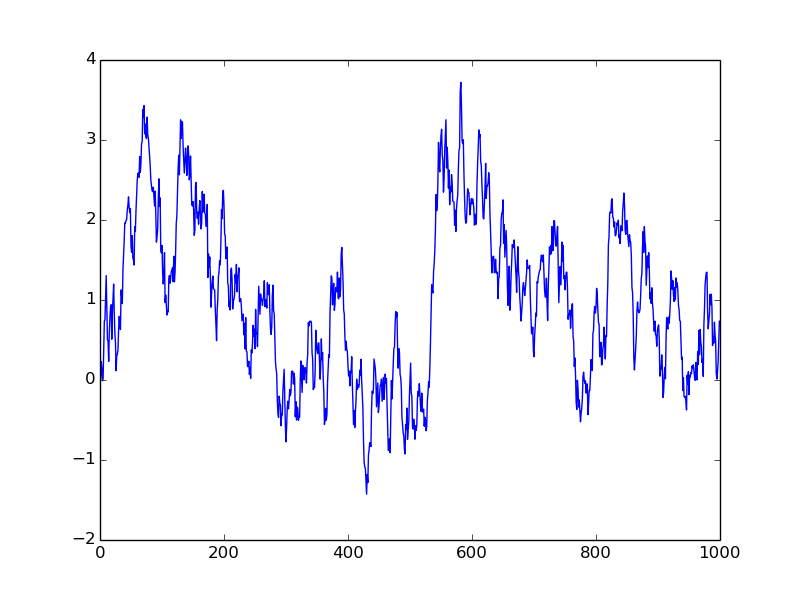}}\\
 Autocorrelations\\  \end{center} }
\parbox{4.5cm}{ \begin{center}
\resizebox{3.5cm}{0.8cm}{\includegraphics{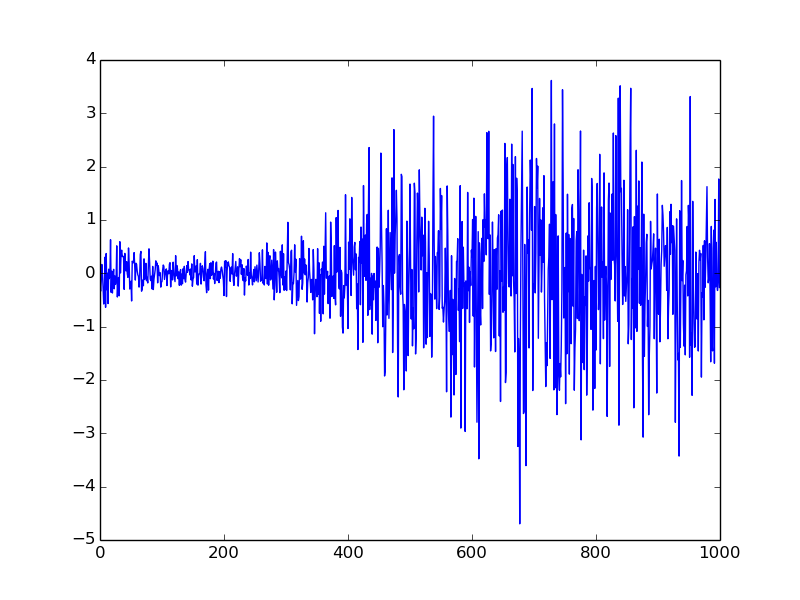}}
\resizebox{3.5cm}{0.8cm}{\includegraphics{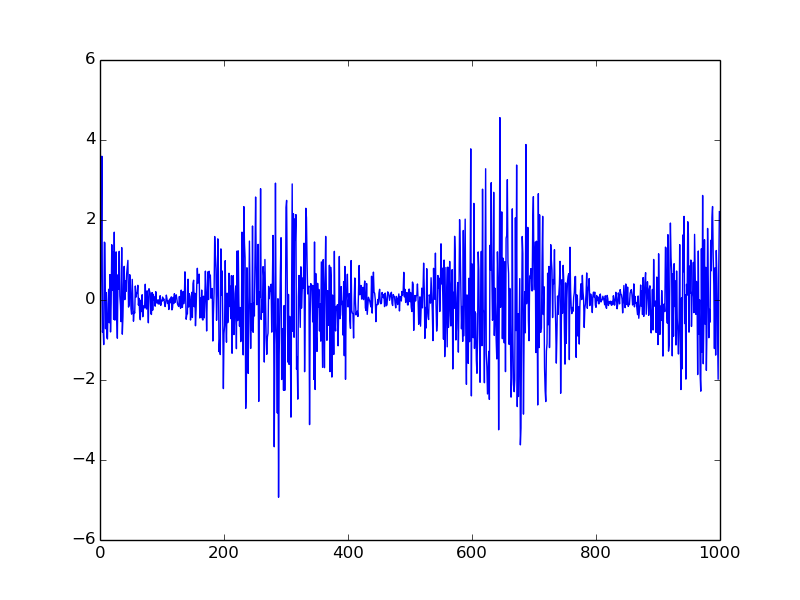}}
\resizebox{3.5cm}{0.8cm}{\includegraphics{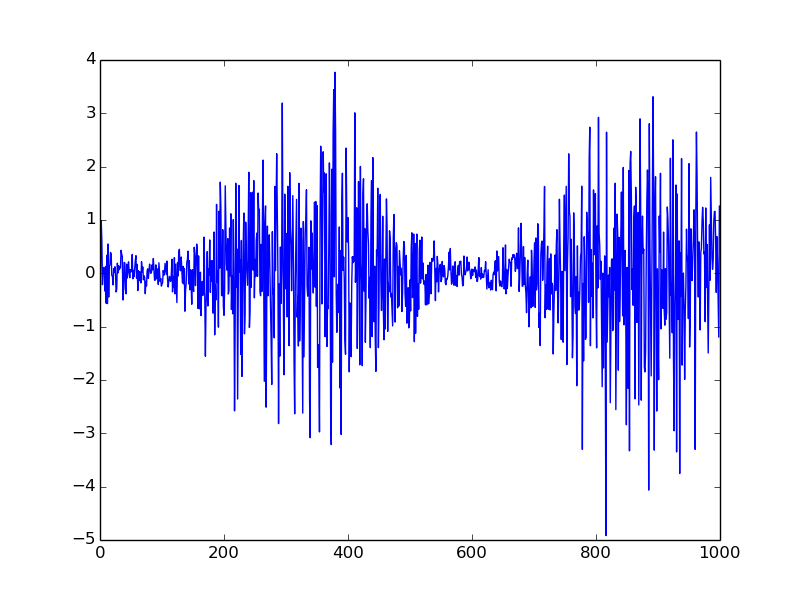}}\\
Nonstationarity\\ \end{center}}
\caption{The two fundamental kinds of temporal structure used in nonlinear ICA. Left: Autocorrelated sources, which show the basic form of temporal dependencies. Right: Nonstationary sources, in particular of exhibiting nonstationarity of variances. If the independent components have either of such structures, the model can be identifiable. Furthermore, such structures can be combined or generalized in various ways.\label{nica.fig}}
\end{figure}

The two kinds of temporal structure used here are illustrated in their basic forms in Fig.~\ref{nica.fig}.
Such a dichotomy of temporal structure is useful for intuition about the temporal dependencies. However, nonstationarity can be subsumed as a case of temporal dependencies by assuming that the apparent nonstationarity stems from a hidden Markov model \citep{Halva20UAI}. This leads to the general theory of \citet{Halva21} which shows identifiability for very general temporal dependencies, including those coming from HMM's. A related model combining the two kinds of temporal structure was proposed by \citet{Morioka21AISTATS}. 

Another way of generalizing identifiability is to realize that the proofs based on nonstationarity simply assume that conditionally on the segment index, the distributions of the components change, while they impose no temporal structure inside the segments. Thus, we can replace this segment index by any observed \textit{auxiliary variable} that conditions the distributions of the components. It could be the index of a subject in a biomedical setting, the index of a country in an international study, or anything that changes the underlying statistics in a suitable way. 
Thus, we get another set of identifiability results, based on observation of such an additional auxiliary information \citep{Hyva19AISTATS}. \citet{Khemakhem20iVAE,Khemakhem20NIPS} generalize those results even further; see also \citet{zimmermann2021contrastive}. We will see below how this is particularly useful in the case of causal discovery.

Another generalization of the results is to consider the noisy mixing model, where noise is added to the mixing as in the classical factor analysis in Eq.~\ref{fa}. Such additive noise has some algorithmic advantages and has actually been widely used in related unsupervised models as well as by \citet{Khemakhem20iVAE}. While its distribution is typically considered known, \citet{Halva21} prove the identifiability of the noise distribution under quite general conditions.

\subsection{Sketch of an identifiability proof}

Next we provide a sketch of a simple identifiability proof following \citet{Hyva16NIPS}. This is only a special case of the currently available theory, but serves to illustrate the basic principles. For the most sophisticated currently available proofs based on \citet{Hyva16NIPS}, the reader is referred to \citet{Khemakhem20iVAE,Khemakhem20NIPS}, whose proofs are thus generalizations of the following. Note also that \citet{Halva21} develop a proof using a very different method, based on \citet{Hyva17AISTATS}, and reminiscent of the linear ICA identifiability proof given above. Another very different approach to the proofs was proposed by \citet{Hyva19AISTATS}.

We consider a nonstationary model where at each segment $\tau=0,\ldots T$, the $i$-th component follows an exponential family of order one:
  \begin{equation} \label{eq:p_tau}
\log p_\t(s_i)=  b_{i}(s_i) +  \lambda_{\t,i} q_i(s_i) 
\end{equation}
with the $q_i,b_i$ being the sufficient statistic and the base measure of component $i$, and $\lambda_{\t,i}$ the parameter for component $i$ and segment $\tau$. (Instead of the segment $\t$, the proof could also be developed for a time index $t$ with no changes.)
Now,  compute the log-pdf of a data point $\x$ in the segment $\t$ under the nonlinear ICA model.  Using the probability transformation formula, the log-pdf is given by
\begin{equation}
    \log p_\t(\x) = \sum_{i=1}^n b_i(g_i(\x)) + \lambda_{\t,i}q_i(g_i(\x)) 
    + \log | \det \ve J \g (\x) | 
   \label{eq:p_x}
\end{equation}
where we drop the index $t$ from $\x$ for simplicity,  and $\g(\x) = (g_1(\x), \ldots , g_n(\x))^T$ is the inverse function of (the true) mixing function $\ve f$; thus, $s_i=g_i(\x)$ by definition.  $\ve J$ denotes the Jacobian matrix.

Assume we now have two different models which give the same data distribution, thus violating identifiability. That is, there is another inverse function $\tilde{\g}$, sufficient statistics and base measure $\tilde{q},\tilde{b}$ and parameters $\tilde{\lambda}$ such that for all $\t$,
\begin{multline}
  \sum_{i=1}^n 
  b_i(g_i(\x)) + \lambda_{\t,i} q_i(g_i(\x))     + \log | \det \ve J \g (\x) | 
  \\=    \sum_{i=1}^n \tilde{b}_i(g_i(\x)) +\tilde{\lambda}_{\t,i}   \tilde{q}_i(\tilde{g}_i(\x))     + \log | \det \ve J \tilde{\g} (\x) | 
\end{multline}
Now, subtract both sides of this equation for the corresponding terms obtained for $\t=0$. We get
\begin{equation} \label{tcleq}
    \sum_{i=1}^n \alpha_{\t,i}q_i(g_i(\x))     
=    \sum_{i=1}^n \tilde{\alpha}_{\t,i}\tilde{q}_i(\tilde{g}_i(\x))   
\end{equation}
with $\alpha_{\t,i}=\lambda_{\t,i}-\lambda_{0,i}$. Remarkably, we got rid of the Jacobian terms.

Now it is enough to collect Eq.~(\ref{tcleq}) for all $\t=1,\ldots,T$. Assume that the matrix of the $\alpha$ is full rank, which means that we have enough segments (their number is greater than the dimension of the data) and there is enough variability in their distributions. Then, we can solve for the $\tilde{q}_i(\tilde{g}_i(\x))$ which are then given  by a linear transformation of the $q_i(g_i(\x))$. Such a linear indeterminacy can be resolved by linear ICA assuming that the components are marginally independent (in addition to conditionally independent in each segment). Thus, we see that $\tilde{q}_i(\tilde{g}_i(\x))=q_j(g_j(\x))=q_j(s_j)$ for some permutation of the indices $i,j$, and we obtain the components up to the pointwise nonlinear transformation given by the sufficient statistics. (The pointwise transformation is characterized in detail by \citet{Khemakhem20iVAE}, who actually don't need a condition of marginal independence.)\qed

\subsection{Alternative approach by constraining nonlinearity}

An alternative approach to making nonlinear ICA identifiable consists of constraining the mixing function. A natural approach would be to assume that the mixing function is close to linear \citep{Zhang08}. Recent research has focused on imposing constraints on the Jacobian matrix $\J\f$ of the mixing function to achieve something to that effect.

The classic Liouville's theorem of conformal mappings is highly relevant here. It considers the case where the Jacobian is almost orthogonal in the sense that
\begin{equation} \label{orth}
  \J\f(\x)^T\J\f(\x)=\alpha(\x)\mathbf{I}
\end{equation}
for a scalar-valued function  $\alpha$. The theorem states that any such sufficiently smooth function $\f$ has to be belong to a very restricted class of functions (called M\"obius transformations). A simple corollary of the theorem, stated and proven in Appendix~\ref{liouville.sec}, says that if $\alpha\equiv 1$, i.e.\ the Jacobian is orthogonal, the function $\f$ is actually necessarily affine: $\f(\x)=\U\x+\b$ for some orthogonal matrix $\U$. This theory thus provides a strong result on what kind of constraints on the Jacobian are meaningful. Constraining the Jacobian to be orthogonal is not meaningful since it does not allow for any nonlinear mixing functions. An intriguing question is, therefore, to explore some relaxations of the constraint in Eq.~(\ref{orth}). Hopefully, some such relaxations would allow for a sufficiently large class of nonlinear functions, while still providing identifiability \citep{gresele2021independent,zimmermann2021contrastive,buchholz2022function}.

A number of further approaches constraining the mixing have recently been proposed. \citet{kivva2022identifiability} show identifiability in the case of a piecewise affine mixing function.  \citet{moran2021identifiable} impose a restriction on how many observed variables are influenced by a single independent component, thus leading to a special kind of ``sparsity'' of the mixing function. 
In a slightly different context, \citet{donoho2003hessian} consider a constraint of orthogonality of the Jacobian (but for non-invertible $\f$) in the case of dimension reduction, which \citet{horan2021unsupervised} apply on nonlinear ICA combined with dimension reduction. Furthermore, \citet{willetts2021dont} automatically learn the auxiliary variables from observations by solving a secondary task, even without temporal structure; while \citet{gresele2020incomplete} use a different ``view'' of the data instead of auxiliary variable.

\section{Nonlinear Structural Equation Model} \label{nonsens.sec}

Just like the theory of linear ICA helped in solving the linear SEM problem in Section~\ref{sem.sec}, it turns out that the theory of nonlinear ICA helps in solving the nonlinear SEM problem. 
In what follows we focus on the bivariate causal discovery problem. This corresponds to recovering the causal structure using observations from two variables, which we denote by 
$x_1$ and $x_2$. While bivariate causal discovery is simplified special case of the 
more general causal discovery problem, it remains a challenging task.

\subsection{Definition of problem}

We start by defining a fully nonlinear SEM of arbitrary dimension as
\begin{equation}
        \label{eq:intro:sem}
        x_j = f_j(\mathbf{PA}_j, e_j), \quad j=1,\ldots,\ddim,
\end{equation}
for arbitrary nonlinear functions $f_i$.
The variables $\mathbf{PA}_j \subseteq \{x_1, \ldots, x_\ddim\} \setminus x_j $ are the parents of the variable $x_j$ in the associated graph. The point is that the ``parents'' cause the variable $x_j$, and in the general case the problem of causal discovery can be framed as finding which variables are parents of which.

In order to accomplish identifiability, causal discovery algorithms generally adopt one
of two techniques.  The first approach is to impose constraints on the
functions $f_j$ that define the SEM~\eqref{eq:intro:sem}. In the extreme case, we obtain linear models, possibly with non-Gaussian external influences as reviewed above. In contrast to the ICA case, however,  some quite meaningful restricted nonlinear models have also been proposed for SEM and will be reviewed below. Nevertheless, arguably the most interesting case is where the nonlinearities are unrestricted while we impose some statistical assumptions on the external influences, which is our main topic here.

\subsection{Identifiable weakly nonlinear causal models}
In this subsection, we will review some of the most notable identifiable
nonlinear causal models based on strong restrictions on the nonlinearity.
\paragraph{Additive noise model (ANM).}
\citet{Hoyer09} introduced the additive noise model, in which the SEM has the form
\begin{equation*}
        x_j = f_j(\mathbf{PA}_j) + e_j,
\end{equation*}
and the noise variables $\mathbf{e}$ are both mutually independent and independent of $\mathbf{x}$.
Their theoretical identifiability result focuses on the case of two variables
$x_1$ and $x_2$.  It stipulates that if $x_1$ causes $x_2$, which we denote by $x_1 \rightarrow x_2$, then we cannot
write $x_1 = g(x_2) + \tilde{e}$ for some function $g$ and noise $\tilde{e}$ that is independent of $x_2$.  Essentially, this SEM is
asymmetrical with respect to $x_1$ and $x_2$ and can only describe the natural
cause-effect relationship. In other words, it is identifiable.
\citet{peters2014causal} generalized the identifiability result to the case of more than two variables.
\paragraph{Post-nonlinear model (PNL).}
\citet{Zhang09UAI} introduced the post-nonlinear model, which
generalizes ANM by adding a subsequent invertible mapping $g_j$:
\begin{equation*}
        x_j = g_j(f_j(\mathbf{PA}_j) + e_j).
\end{equation*}
The noise variables $\mathbf{e}$ are still assumed to be mutually independent
and independent of the causes.  The authors show that the bivariate PNL model
is identifiable in most cases and enumerate five special situations in which
the model is not identifiable.  This identifiability theory generalizes that of
ANM, which is a special case when $g_j$ is the identity mapping.  Note that if we knew $g_j$, we could reduce the PNL model to an ANM by transforming the effect through the
inverse of the mapping $g_j$, the transformed variable $g_j^{-1}(x_j)$ being
a deterministic function of the original effect $x_j$.

\paragraph{Causal autoregressive flow (CAREFL).} \citet{Khemakhem21AISTATS} note that SEMs are closely related to a class of model called (affine autoregressive) normalizing flows
\citep{rezende2015variational,huang2018neural} in machine learning.
That theory leads to the following definition of a SEM on the observations $\x$:
\begin{equation}
\label{eq:carefl:flow_def_x_ns}
    x_j = e^{\alpha_j(\mathbf{PA}_j)}z_j + \beta_j(\mathbf{PA}_j),\quad j=1,2
\end{equation}
where $z_1,z_2$ are statistically independent latent noise variables, and $\alpha_j(\mathbf{PA}_j)$ and $\beta_j(\mathbf{PA}_j)$ are scalar-valued functions, defined constant (with respect to $\x$) when there are no parents. 
This affine causal model generalizes the additive noise model \citep{Hoyer09}
by adding a cause-dependent coefficient to the noise variable $z_j$ in the SEM; thus, the disturbance $e^{\alpha_j(\mathbf{PA}_j)}z_j$  is not independent of the cause $\beta_j(\mathbf{PA}_j)$, in contrast to most models. (We use here the notation by \citet{Khemakhem21AISTATS} where $z$ is used instead of $e$, the two terms on the RHS are switched, and Greek letters are used for the functions; but the real difference to ANM is in the modulation of the noise.)
Identifiability of the model is shown by  \citet{Khemakhem21AISTATS} under very general conditions. Recent work by \citet{immer2022identifiability,strobl2022identifying} has modified and generalized the identifiability conditions.

\subsection{Identifiability theory for general nonlinearities}

Next we describe a method for estimating a nonlinear SEM for general nonlinearities; this leads to an identifiability result as well. 
We assume we observe bivariate data 
$\textbf{x}(\myindex) \in \mathbb{R}^2$
where $\myindex$ provides an index over all observations ($\myindex$ may be, e.g., a time index  but this is not necessary).

Importantly, we further assume 
data is available over a set of 
distinct environmental conditions or segments, $C \in \mathcal{C}$.  %
As such, each $\textbf{x}(\myindex)$ is allocated to one such environmental condition %
$C(t) \in \mathcal{C}$.
In the case of causal discovery, such environmental conditions could be due to different interventions or due to some changes in the internal dynamics of the system being observed. %
To clearly align the proposed method with the terminology of nonlinear ICA reviewed in 
Section \ref{nica.sec}, 
we note that we may consider each environmental condition as a distinct segment in nonstationarity-based nonlinear ICA. As noted above, the segment should be interpreted in a very general sense, as in the theory of nonlinear ICA by auxiliary variables \citep{Hyva19AISTATS,Khemakhem20iVAE}.

\begin{figure}%
	\begin{center}
		\resizebox{0.7\textwidth}{!}{\includegraphics{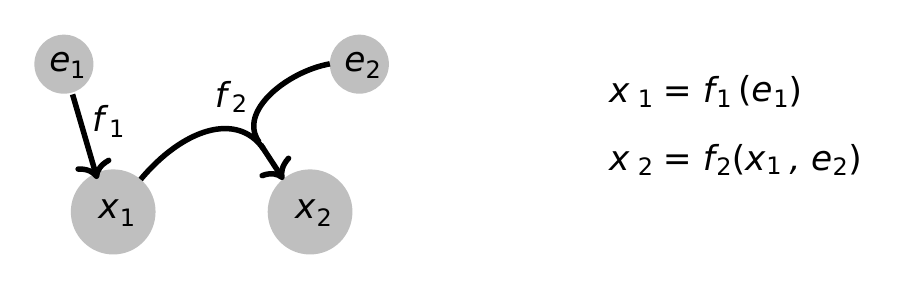}}
	\end{center}
	\caption{Visualization of DAG, $\mathcal{G}$, associated with the SEM
		in equations (\ref{bivariate_eq1}) and (\ref{bivariate_eq2}). 
		The associated structural equations are provided on the right. Note that in contrast to Fig.~\ref{dag.fig}, the disturbance variables are shown as well.}
	\label{DAG_Fig}
\end{figure}

Next we outline the method for causal discovery over bivariate data by \citet{Monti19UAI}, which they termed
{Non}-linear {S}EM {E}stimation using {N}on-{S}tationarity (NonSENS).
Without loss of generality, we explain the basic logic assuming that ${x}_1 \rightarrow {x}_2$, such that 
the associated SEM is of the form:
\begin{align}
\label{bivariate_eq1}
{x}_1(\myindex) &= f_1( e_1(\myindex) ),\\
{x}_2(\myindex) &= f_2({x}_1(\myindex), e_2(\myindex)),
\label{bivariate_eq2}
\end{align}
where $e_1, e_2$ are %
latent disturbances whose distributions are
assumed to vary across environmental conditions, thus creating the non-stationarity. The DAG associated with Equations (\ref{bivariate_eq1}) and (\ref{bivariate_eq2}) is shown in Figure \ref{DAG_Fig}.
The key idea here is that the latent disturbances in a bivariate SEM, $\textbf{e}$, can be interpreted as corresponding to the independent 
sources in a non-linear ICA model, $\textbf{s}$, not unlike in the original LiNGAM theory reviewed above based on \citep{Shimizu06JMLR}.

The proposed NonSENS algorithm consists of a two-step procedure.
First, it seeks to recover latent disturbances 
via non-stationarity-based nonlinear ICA. We note that the non-stationarity introduced by the various environmental conditions, $C \in \mathcal{C}$, 
implies that the method by \citet{Hyva16NIPS} is well suited to recover the source variables $e_1$ and $e_2$. 
Given estimated components, 
we may employ knowledge regarding the statistical independences between 
observed data and estimated sources in order to infer the causal structure, based on
an interesting independence property of the components pointed out by \citet{Monti19UAI}. Denote by $ x \bigCI y$ statistical independence of $x$ and $y$, and by $\nbigCI$ lack of independence. We have:
\begin{proposition}
	\label{Prop_DAG_indep} Assume the true causal structure follows equations  (\ref{bivariate_eq1}) and (\ref{bivariate_eq2}), as depicted in 
	Figure \ref{DAG_Fig}. Then it follows that $x_1 \bigCI  e_2 $ while %
	$ x_1 \nbigCI e_1$ and $ x_2 \nbigCI e_1$ as well as $ x_2 \nbigCI e_2$.
\end{proposition}

This proposition highlights the relationship between observations $\textbf{x}$
and true latent sources, $\textbf{e}$: The cause is independent of one of the disturbances, while the effect is not independent of any disturbance.
In practice, nonlinear ICA returns estimated 
latent sources whose ordering is random. As a result,  we must 
test for independence between each of the variables, $x_1$ and $x_2$, and 
each of the estimated latent sources, $\hat e_1$ and $\hat e_2$. This results in a total of
four distinct independence tests.

The proposed method is thus recapitulated as follows: After estimating %
	(point-wise transformations of) latent variables by nonlinear ICA,
	we make the four statistical independence tests, and conclude a causal 
	effect in the case where, for one of the directions, there is evidence to reject the null hypothesis of independence in three of the tests and 
	only one of the tests fails to reject the null. In such a case, 
the observed variable
regarding which one null hypothesis was not rejected is identified as 
the cause variable.

We finally note that the proposition and method just described based on \citet{Monti19UAI} constitute, in fact, a constructive identifiability proof: The direction of effect in arbitrary nonlinear SEM is identifiable, essentially assuming the data is divided into some conditions that fulfill the assumptions of nonlinear ICA methods based on nonstationarity. 

\section{Discussion}\label{disc.sec}

In this section, we discuss the actual utility of identifiability in applications, some questions for future research, as well as the question of algorithm development.

\subsection{Utility of identifiability}

\subsubsection{Interpretability; finding causal direction} The utility of identifiability is obvious in the case where the parameters (or latent variables) of the model give directly some useful information about the phenomenon being analyzed. This is very clear in the case of causal analysis, where the whole point is to find out which variable causes which, and if the model is not identifiable, the analysis is typically not possible at all. Likewise, in the case of linear ICA, the components obtained often correspond to meaningful phenomena, as in the case of blind source separation (Fig.\ref{icaillufig}). This has been extensively used in neuroscience, for example, where the sources may correspond to sources of activity in the brain \citep{Hyva10NI}.

\subsubsection{Feature extraction; semi-supervised and transfer learning} What is less understood is the utility of identifiability in nonlinear ICA. In deep learning, the components are often very difficult to interpret, since the nonlinearities implemented by the neural networks are difficult to understand or visualize. Therefore, the main utility of identifiability of nonlinear ICA might be found elsewhere. In fact, a representation learned by unsupervised deep learning is often used for some further purpose; for example, the features are used in a further classification task. So, the question is whether the identifiability of the components is useful for such a task.

To begin with, we note that even if the computation of the components is not possible to interpret, single components might still be useful for classification and decision-making. For example, single components might work as biomarkers in a biomedical setting, and finding such biomarkers clearly requires identifiability. 
In a more general setting, \citet{Zhu23} show that nonlinear ICA is very useful for analysing neuroscience data in a \textit{semi-supervised setting}, which means that first the components are learned from a big generic data set, and then used on a smaller data set to solve a specific classification task using a simple linear classifier. In a similar vein, \citet{Khemakhem20NIPS} propose some theory on how identifiable representations would be particularly useful for \textit{transfer learning}, which is a related task where we have data from many different data sets (e.g.\ subjects in a biomedical setting) and then want to generalize to new subjects. 
While these developments point out the empirical utility of nonlinear ICA, they don't quite conclusively prove the utility of identifiability of nonlinear ICA; more developments on that topic would be warranted.

\subsection{Questions for future research}

\subsubsection{Combining causal analysis with feature learning} In this paper, the methods are either finding some hidden factors or doing causal discovery. A very interesting topic would be to combine the two, so that the model learns  hidden factors and causal connections \textit{between} the hidden factors. This is what some would call ``causal representation learning''. In the linear case, such methods can be found in \citet{Zhang09UAI,Monti18UAI}. In the nonlinear case, this is a topic of great current interest in deep learning, see e.g.\  \citet{lachapelle2022disentanglement,Morioka23AISTATS} for some developments. 

\subsubsection{Dependent components; causal discovery with confounders}
In fact, models with causal dependencies between the latent variables are a special case of models where the latent variables are not independent. A general framework for this case was proposed by \citet{Khemakhem20NIPS}, who developed an extension of nonlinear ICA, called Independently Modulated Component Analysis (IMCA), where the components are allowed to dependent. They key idea is that the dependencies are assumed to be stationary, or independent of the auxiliary variable. Allowing such dependencies is likely to be useful in many contexts, and another highly promising topic for future research.

Another potential application of the IMCA framework in causal discovery is to allow for the presence of confounding. As already mentioned, a confounder is a hidden variable that affects both dependent and independent variables, resulting in spurious associations in the causal graph.
We can establish a relationship between IMCA and confounded structural equation models (SEM) since the dependencies of the disturbances can be considered to be induced by the unobserved confounders.
The identifiability of IMCA implies that the causal direction of such an SEM is likewise identifiable.
However, because it is based on independence tests, the estimation technique of NonSENS based on testing described above cannot be used here.
Possibly, a non-constraint-based method, such as likelihood ratio measures \citep{Hyva13JMLR,Monti19UAI}, might still be pursued.

\subsubsection{Identifiability of intermediate layers}
If the nonlinear function in nonlinear ICA is modelled by a neural network, we are estimating much more than only a single nonlinearity. In fact, the intermediate layers in neural networks are frequently used as useful features for a later classification task.
They may even be preferable in some applications over the representations learned by the final layer \citep{alain2018understanding,chen2020simple,mikolov2013efficient}.
An intriguing question arises:
can the identifiability results of the representations learnt by nonlinear ICA be generalized to previous layers?
\citet{Khemakhem20NIPS} showed that some such neural network architectures are, in fact, identifiable, using a  form of induction to ``propagate identifiability'' forward through the network.
Thus, a potential avenue of research is to prove that the intermediate layers preceding a final  layers in a neural network can inherit the identifiability guarantees of the final layer.
Extending such proofs to convolutional networks would be particularly interesting since they are frequently utilized in image learning and have a strong mathematical theory \citep{wiatowski2017mathematical}.

\subsection{Estimation algorithms}
Finally, let us very briefly discuss what kind of estimation methods are available.
Basically, we only need methods to estimate ICA, including nonlinear ICA,  since the estimation of SEMs can be reduced to ICA. In machine learning, we typically distinguish between two parts of an estimation method: 1) an estimation principle (such as maximum likelihood), typically leading to an objective function, and 2) a computational algorithm, typically optimizing the objective function. For linear ICA, we proposed FastICA \citep{Hyva99TNN} which is widely used and was originally also used to solve the SEM estimation by \citet{Shimizu06JMLR}, although further methods for SEM were subsequently developed e.g.\ by \citet{Shimizu11JMLR,Hyva13JMLR}.

For nonlinear ICA, we have made several proposals; for an in-depth exposition, see \citet{Hyva23pattern}. First, we proposed ``self-supervised'' methods \citep{Hyva16NIPS,Hyva17AISTATS,Hyva19AISTATS}, which is a class of methods of great current interest in machine learning. Subsequently, we proposed maximum likelihood estimation, starting with models with additive noise \citep{Khemakhem20iVAE,Halva21} that enable variational approximation. The case of maximum likelihood estimation in the basic noise-free model used in this paper initially looks easy, but presents serious computational problems, which we have largely solved in \citet{Gresele20}. In any case, developing better algorithms is definitely an interesting topic for future research as well.

Once we have an estimating algorithm, the question of its finite-sample performance can be considered. For linear ICA, the asymptotic variance (statistical efficiency) has been analyzed by several authors, including \citet{Cardoso96,Pham97,Hyva97NNSP,tichavsky2006performance}, and these results may be more or less directly applicable to linear SEM estimation as well. Some analysis of the robustness against outliers has also been performed, in both linear and nonlinear cases \citep{Hyva97NNSP,Sasaki20UAI}. For the nonlinear case, we are not aware of any analysis of statistical efficiency except in terms of simulations. These are clearly interesting points for future research.

\section{Conclusion}\label{conc.sec}

We started this review from the linear, Gaussian factor analysis model which is known to be unidentifiable since the 1950's if not earlier. The identifiability problem was solved by independent component analysis, a \textit{non}-Gaussian factor analysis model which is identifiable in the precise sense that it can recover components that actually created the data, as in the case of blind source separation. Linear ICA further enables estimation of linear structural equation models, thus leading to identifiable causal discovery.
ICA can be made nonlinear, but this is not at all straightforward: new assumptions are needed for identifiability. Here, we considered mainly temporal dependencies and nonstationarity (the latter being  taken in a very general sense). Restricting the nonlinear mixing is also likely to work but few successful models have been proposed so far. As in the linear case, we finally saw that nonlinear ICA enables identifiability and estimation of nonlinear SEM as well.

\appendix
\section{Liouville's theorem and nonlinear ICA} \label{liouville.sec}

A possible line of research for making nonlinear ICA identifiable is imposing suitable conditions on the Jacobian of the transformation. The conditions are typically related to orthogonality of the Jacobian, also called local isometry \citep{gresele2021independent,zimmermann2021contrastive,buchholz2022function}. In this appendix,  we consider the set of invertible mappings within a real space, i.e. such that the dimension is not changed, as is typical in ICA theory. We show that functions which are locally isometric and $\mathbb{R}^n \rightarrow \mathbb{R}^n$  are necessarily affine, and thus do not provide a meaningful basis for nonlinear ICA theory. We do this by proving a variant of the theorem on conformal mappings by Liouville. While the result can be considered well-known, we provide a very simple proof for this variant, which is difficult to find in the literature.

\subsection{Definitions}

We start by some definitions:

\begin{definition}\label{def:conformal}
  Let  $\f$ be a differentiable mapping from an open subset $U$ of $\mathbb{R}^n$ to an open subset $V$ of $\mathbb{R}^n$.
  The mapping called \textbf{conformal} if
  \begin{equation}
    \J\f(\x)^T    \J\f(\x)=c(\x) \mathbf{I}
  \end{equation}
  where $\J\f$ is the Jacobian matrix (of partial derivatives) of $\f$, and $c(\x)$ takes positive scalar values. %
\end{definition}
The geometrical meaning of the definition is that the mapping preserves angles, but the scaling can be changed by $c(\x)$.
An important special case is obtained when the Jacobian is orthogonal:
\begin{definition}\label{def:isometric}
  If in Def.~\ref{def:conformal}, $c(\x)\equiv 1$, the mapping is called \textbf{locally isometric}.
\end{definition}
In a geometrical interpretation, the mapping is required to preserve both angles and the scaling. This is related to the definition used by \citet{donoho2003hessian}, and based on them, by \citet{horan2021unsupervised}. However, those authors consider the case where the dimension is reduced by the mapping $\f$, and thus our results are rather different. In fact, local isometry is typically defined in the context of manifolds which are of a lower dimension than the space itself, and therefore our definition is a special case of the more conventional one. The orthogonality of the Jacobian has also been used, more heuristically, as a regularizer \citep{wei2021orthogonal,kumar2020implicit}.

Merely for simplicity, we further introduce the following terminology:
\begin{definition}\label{def:orthaff}
  A function $\f$ is called \textbf{orthogonally affine} if it is of the form
  \begin{equation} \label{orthogonal}
    \f(\x)=\U\x+\b
  \end{equation}
  for some constant vector $\b$ and an orthogonal matrix $\U$.
\end{definition}
Such a function is also called ``rigid motion'' in some contexts.

\subsection{Locally isometric functions are orthogonally affine}

  Our analysis is based on a well-known theorem on conformal mappings by \mbox{Liouville} from 1850. 
  To keep the presentation short, we simply provide our variant of the theorem, and refer the reader to the literature for the original theorem \citep{flanders1966liouville,nevanlinna1960differentiable}.   Our variant of the theorem is as follows:
\begin{theorem} \label{theorem}
  Assume $\f$ is locally isometric (Def.~\ref{def:isometric}), and in $\Ctwo$ (i.e.\ with two continuous derivatives),  in a real space of dimension $n\geq 2$. 
  Then, $\f$ is orthogonally affine (Def.~\ref{def:orthaff}).
\end{theorem}
\textit{Proof:}
This proof is closely related to the first half of the proof of Liouville's theorem by \citet{flanders1966liouville}; see \citet{nevanlinna1960differentiable} for another well-known proof. 

We drop the argument $\x$ for notational simplicity, and denote by $\w\cdot\v$ the dot-product. Denote by $\J_i$ the $i$-th column of the Jacobian of $\f$, i.e.\ the vector of the partial derivatives with respect to $x_i$. Consider the $i,j$-th element of the matrix equation defining isometry by orthogonality of the Jacobian:
  \begin{equation} \label{isometryinproof}
 \J_i \cdot \J_j=\delta_{i=j}
  \end{equation}
  Take the derivative with respect to $x_k$ of both sides:
  \begin{equation} 
   \J_i\cdot \J_j^k + \J_j\cdot \J_i^k=  0
  \end{equation}
  where $\J_i^k$ is the vector of the partial derivatives with respect to $x_k$ of the entries of $\J_i$.
  Thus, we get the following skew-symmetricity condition:
  \begin{equation} \label{skewsymm}
   \J_i\cdot \J_j^k  = - \J_j\cdot \J_i^k
  \end{equation}
On the other hand, by the fact that the order of derivation can be changed, we have the symmetricity condition:
  \begin{equation} \label{exchange}
   \J_i\cdot \J_j^k  =  \J_i\cdot \J_k^j
 \end{equation}
Obviously, a matrix which is both symmetric and skew-symmetric is necessarily zero. Here we actually have a tensor but we show next that the same principle applies to such a tensor; this result is sometimes known as the Braid Lemma.
Take any indices $i,j,k$, and apply (\ref{skewsymm}) and (\ref{exchange}) in alternation, three times. We get
\begin{equation}
   \J_i\cdot \J_j^k  = - \J_j\cdot \J_i^k = - \J_j\cdot \J_k^i = \J_k\cdot \J_j^i  =\J_k\cdot \J_i^j  = - \J_i\cdot \J_k^j = - \J_i\cdot \J_j^k 
\end{equation}
The equality of the first and last term shows that they must be zero. This holds for all $i$, and the $\J_i$ form an orthogonal basis, which implies that $\J_j^k$ is zero. Thus, all second-order partial derivatives vanish at every point, and $\f$ must be affine. By definition of isometry, it must further be orthogonally affine. \qed

Our theorem extends the Liouville theory in the sense that our theorem applies to $n\geq 2$ while Liouville's theorem assumes $n\geq 3$. On the other hand, ours is a special case since Liouville assumes conformality while we assume local isometry.  Given Liouville's theorem it would be very easy to prove a corollary which gives the desired result for $n\geq 3$ but we prefer to prove a theorem which is not a strict corollary but an extension at the same time, without any help from such advanced theory.  It was in fact possible to give a very simple proof in our case, while only the very complicated proofs seem to be available for Liouville's theorem.

The implication is that a nonlinear ICA model (or any deep latent variable model) where the (dimension-preserving) mixing is assumed locally isometric trivially reduces to a linear ICA model. (Note that we didn't use any statistical properties of any components here.) The assumption of local isometry is far too strong from this viewpoint. 

\end{document}